\documentclass{article}

\usepackage[final,nopatch=footnote]{microtype}

\usepackage{graphicx}
\usepackage{subfigure}
\usepackage{booktabs} 

\usepackage{hyperref}

\usepackage{amsmath,amsthm, graphicx, multicol, array, comment, caption, rotating, longtable, xpatch, bbm, color, float,setspace, bm, enumerate,blkarray,arydshln,multirow}

\newcommand{\bbR}{\mathbb{R}}
\newcommand{\bbE}{\mathbb{E}}

\newcommand{\calN}{\mathcal{N}}
\newcommand{\calX}{\mathcal{X}}

\newcommand{\bb}[1]{\left[#1\right]}
\newcommand{\bp}[1]{\left(#1\right)}
\newcommand{\bc}[1]{\left\{#1\right\}}

\theoremstyle{plain}
\newtheorem{proposition}{Proposition}

\newtheorem{lemma}{Lemma}
\newtheorem{theorem}{Theorem}
\newtheorem{corollary}[theorem]{Corollary}

\newtheorem{assumption}{Assumption}
\newtheorem{remark}{Remark}

\usepackage[dvipsnames]{xcolor}
\definecolor{tencent_blue}{RGB}{0, 82, 217}
\definecolor{tencent_orange}{RGB}{238, 126, 71}
\usepackage[accepted]{icml2024}

\usepackage{amsmath}
\usepackage{amssymb}
\usepackage{mathtools}
\usepackage{amsthm}
\usepackage{graphicx}
\usepackage{epstopdf}

\usepackage{color-edits}
\definecolor{redorange}{RGB}{255, 68, 51}

\usepackage[capitalize,noabbrev]{cleveref}

\theoremstyle{plain}

\usepackage[textsize=tiny]{todonotes}

\icmltitlerunning{Statistical Properties of Robust Satisficing}

\begin{document}

\twocolumn[
\icmltitle{Statistical Properties of Robust Satisficing}

\begin{icmlauthorlist}
\icmlauthor{Zhiyi Li}{sch1}
\icmlauthor{Yunbei Xu}{sch2}
\icmlauthor{Ruohan Zhan}{sch3,SHCIRI}
\end{icmlauthorlist}

\icmlaffiliation{sch1}{School of Mathematical Sciences, Peking University, 
Beijing, China}
\icmlaffiliation{sch2}{Yunbei Xu, Laboratory for Information and Decision Systems, Massachusetts Institute of Technology, Cambridge, MA, USA}
\icmlaffiliation{sch3}{Department of Industrial Engineering and Decision Analytics, The Hong Kong University of Science and Technology, Clear Water Bay, Kowloon, Hong Kong SAR}
\icmlaffiliation{SHCIRI}{HKUST Shenzhen-Hong Kong Collaborative Innovation Research Institute, Futian, Shenzhen, China}

\icmlcorrespondingauthor{Ruohan Zhan}{rhzhan@ust.hk}

\icmlkeywords{Machine Learning, ICML}

\vskip 0.3in
]

\printAffiliationsAndNotice{}   
\begin{abstract}
The Robust Satisficing (RS) model is an emerging approach to robust optimization, offering streamlined procedures and robust generalization across various applications.  
However, the statistical theory of RS remains  unexplored in the literature. This paper fills in the gap by comprehensively    analyzing the theoretical properties of the RS model.
Notably, the RS structure offers a more straightforward path to deriving statistical guarantees compared to the seminal Distributionally Robust Optimization (DRO), resulting in a richer set of results. 
In particular, we establish two-sided confidence intervals for the optimal loss without the need to solve a minimax optimization problem explicitly.  We further provide  finite-sample generalization error bounds for the RS optimizer. 
Importantly, our results extend to scenarios involving distribution shifts, where discrepancies exist between the sampling and target distributions. 
Our numerical experiments show that the RS model consistently outperforms the baseline empirical risk minimization in small-sample regimes and under distribution shifts. Furthermore, compared to the DRO model, the RS model exhibits lower sensitivity to hyperparameter tuning, highlighting its practicability for robustness considerations.

\end{abstract}

\renewcommand\labelitemi{\ensuremath{\bullet}}
\newcommand\Bernoulli{\operatorname{Bernoulli}}
\newcommand\Dirichlet{\operatorname{Dirichlet}}
\newcommand\df{$\Delta$-fit\xspace}
\newcommand\supp{\operatorname{supp}}
\newcommand\sign{\operatorname{sign}}
\newcommand\cX{\mathcal{X}}
\newcommand\wh{\widehat}
\newcommand\Ball{\operatorname{B}}
\newcommand\RE{\operatorname{RE}}
\newcommand\risk{\mathcal{R}_{0/1}}
\newcommand\mse{\mathcal{R}_{\operatorname{sq}}}
\newcommand\cH{\mathcal{H}}
\newcommand\sgn{\operatorname{sgn}}
\newcommand\DJH[1]{{\color{red}[\textbf{DJH}: #1]}}

\section{Introduction}
Robust methods are optimization techniques that guarantee performances  even when environments vary slightly \cite{ronchetti2021main}.  These methods are resilient against variations or uncertainties, ensuring consistent and reliable outcomes. Robustness provided by these methods is particularly valuable in  scenarios where limited sample sizes may not fully capture  the entire distribution, or where the target environment differs from the initial sampling distribution. 

The application of robust methods spans across various domains: in machine learning,  they are utilized to enhance the robustness of algorithms, ensuring they maintain strong performance even when there are adversarial attacks in the  input data  \cite{blanchet2019robust,sim2021tractable}. In energy systems, they are adopted to optimize the operation and planning, including bidding strategies in electricity markets, operation scheduling of power systems, and integration of renewable energy \cite{li2023data,huang2023distributionally}. In supply chains,  they  are employed to optimize various aspects such as production planning, inventory management, logistics, and transportation. \cite{chen2023designing,deng2023distributionally,wang2023risk}.   These examples represent a fraction of the wide-ranging applications of robust methods. In fact, robust methods can be applied to any field that involves optimization problems, making it a  vital tool for decision-making  under uncertainty. 

Among  various robust methods, Distributionally Robust Optimization (DRO) is a pivotal approach  \cite{hu2013kullback,bayraksan2015data,esfahani2015data}. DRO's significance lies in its more  robust handling of ambiguity compared to conventional stochastic programming models. This is achieved by optimizing the worst-case performance over a set of potential distributions rather than for a single distribution. Specifically, the DRO problem is formulated as follows:

\begin{align}  
&\quad \min_{x\in\mathcal{X}}\max_{P\in \mathbb{P}_r} ~\bbE_P[h(x,\xi)]\label{DRO},
\\
&\mbox{where }\mathbb{P}_r=\{P\in \mathbb{P}:d(P,\hat{P}_N)\leq r\}.\nonumber
\end{align}
Above, $x$ represents the decision variable, which is contained in a non-empty decision space $\mathcal{X}$, and $\xi$ is a random variable. 
The function $h(x,\xi)$ denotes the loss associated with $x$ and $\xi$. $\hat{P}_N$ is the empirical distribution derived from the data\footnote{
Other nominal distributions are also viable. For example, when provided with a parametric distributional class, the distribution estimated using maximum likelihood estimation  can serve as a substitute for $\hat{P}_N$.}.
The function $d(\cdot,\cdot)$ is  a distance measure to quantify discrepancies between distributions. The hyperparameter $r$, referred to as the ``radius'', defines the ambiguity set $\mathbb{P}_r$,  a subset of $\mathbb{P}$ that encompasses all feasible distributions. It plays a crucial role in controlling  robustness—the larger the value of $r$, the greater the robustness demanded.

Despite its strengths,  DRO  has a few shortcomings. 
First, DRO can be overconservative in practice, as \citeauthor{esfahani2015data} pointed out.
This is because the DRO framework optimizes the worst-case scenario in the distribution domain, which may be unnecessarily large to incorporate the target distribution.
Second, selecting an appropriate and interpretable radius $r$ is a challenging task in practice, as noted by \citeauthor{sim2021tractable}. This difficulty stems from the abstract nature of  the radius, which characterizes the  distance within the distributional space and is hard to be intuitively translated into tangible, real-world values.  
In addition,  there is a growing demand for incorporating globalized distributions--as opposed to restricting to an ambiguity set under the DRO framework--to further increase robustness \cite{liu2023globalized}.

To address these issues, the Robust Satisficing (RS) model has been proposed \cite{long2023robust}, as structured below:
\begin{align}
k_{\tau}=&\min \quad k \label{eq:rs}\\ \text { s.t. } & \bbE_{P}[h(x, \xi)]-\tau \leq k d(P,\hat{P}_N), \quad \forall P \in\mathbb{P} \nonumber \\ & \boldsymbol{x} \in \mathcal{X}, \quad k \geq 0. \nonumber
\end{align}
Here, the hyperparameter is no longer the radius $r$ of the ambiguity set, but a reference value $\tau$, which can be interpreted as an anticipated cost in practical applications. The constraint \eqref{eq:rs} then ensures that once the expected loss under a certain distribution exceeds our anticipated cost, the excess part should not be too large: it will be controlled by a multiple of the distribution's distance to the empirical distribution $\hat{P}_N$. 
Hence, the RS model compromises some training set performance for robustness in the target distribution, as it doesn't aim for minimizing the empirical loss. 
Unlike DRO that focuses on worst-case optimization, RS follows a satisficing strategy to avoid over-conservatism, thereby providing better generalization performance on the target distribution.
Another key aspect of the RS model is its global consideration of probability distributions, unlike DRO's restriction to ambiguity set.

Current research on the RS model primarly centers around forming tractable new optimization models and experimental analysis.  Notable examples of tractable RS model encompass Risk-based Linear Optimization and Linear Optimization with Recourse \cite{sim2023analytics,long2023robust}, illustrating RS's practical optimization and generalization advantages compared to the DRO models.  \citeauthor{ruan2023robust} proposed Robust Satisficing Markov Decision Processes and demonstrated  its  superiority over  traditional robust MDP through experiments. \citeauthor{saday2023robust} proposed the Robust Bayesian Satisficing model,  established upper bound on regret and outperformed Distributionally Robust Bayesian Optimization in experiments.
Despite RS's notable results in the realm of optimization, to the best of our knowledge, there are no existing studies on the statistical properties of the RS model. This gap leads to  the central research questions of this paper:

 {\it \centering Are there statistical guarantees for the RS model?
What are the statistical merits it holds, potentially surpassing DRO?} 

\subsection{Contributions}

 Our work delves into the statistical theory of the RS model, with a focus on deriving and analyzing its statistical properties. In particular,  we provide a two-sided confidence interval estimate for the  optimal loss using the reference value, and  present non-asymptotic upper bound of the generalization error. These results fill a crucial gap in the literature, where statistical guarantees for the RS model have  been seldom studied.  
  It is noteworthy that our results extend beyond cases where the sampling distribution matches the target distribution of interest, a context where robustness still remains relevant due to potential discrepancies between the empirical and sampling distributions, especially in small-sample regimes;   
 we also consider scenarios involving distribution shifts, where disparities exist between the sampling distribution and the target distribution.  
We highlight the contributions of this paper as follows:

\begin{enumerate}[1)]

\item  We obtain  two-sided, non-asymptotic confidence intervals for  the optimal loss $J^*$ in the RS model,  where $J^*$ is the minimum expected loss under the true distribution. Notably, this result  does not necessitate solving a minimax optimization problem  explicitly. 

\item  We present finite-sample generalization error bounds for the optimizer derived from the RS model, achieved through an insightful and succinct derivation.

\item We demonstrate that, even under distribution shifts, our key findings -- confidence intervals and generalization error bounds for the RS model optimizer -- remain valid. These results incorporate an additional term, a finite multiple of the distance between the sampling and target distributions. This adaptation highlights the RS model's robust generalization abilities.

\item Our numerical experiments reveal that RS model's advantages  over the empirical risk minimization baseline becomes more pronounced in small-sample regimes or with increasing distribution shifts. Furthermore, our analysis reveals the relationship between the RS and DRO models under the Lipschitz loss scenarios, which also highlights that the RS model has lower sensitivity to hyperparameter tuning as compared to DRO.

\end{enumerate} 
In all these aspects, we perform an extensive comparison with DRO. Our analysis reveals that these advantageous properties are closely associated with the inherent structure of the RS model itself. It becomes evident that obtaining statistical guarantees is more straightforward within the RS framework compared to the DRO framework.

\section{Set up}
We start by describing our learning problem. Let $\xi\in\Xi$ be the  $m$-dimensional random variable of observations and $x\in\calX$ be the decision variable to be learned. Let $h(x,\xi)$ be the loss function (which can accommodate a wide range of machine learning problems as detailed in  \cref{sec:ml_loss}).  
We use $J^*$ to denote the minimum expected loss under the optimal decision variable  $x^*$:
\begin{align}\label{J^*}
J^*:=\inf_{x\in\mathcal{X}}\bbE_{P^*}[h(x,\xi)]=\bbE_{P^*}[h(x^*,\xi)].
\end{align}
Given $N$ observations $\{\xi_i\}_{i=1}^N$ sampled from the  distribution $P^*$, the decision maker wants to learn a decision variable  such that the expected loss is minimized.

Consider the \textbf{Robust Satisficing} (RS) model  \eqref{eq:rs}, and  we focus on  the Wasserstein distance for the distance measure between distributions.
Here, $\tau$ is the ``reference value'', which can be interpreted as the anticipated cost in practical applications, and its choice will be further discussed  in Section \ref{sec:reference_value}.  $\mathbb{P}$ is the set of all feasible distributions, on which the RS model does not impose any constraint; this allows   the RS model to consider probability distributions globally. 
$\hat{P}_N$ denotes the empirical distribution of samples $\{\xi_i\}_{i=1}^N$, which converges to $P^*$ as sample size goes to infinity. And $d_{\mathrm{W}}$ denotes the type-1 Wasserstein distance between two distributions\footnote{We follow the literature \cite{long2023robust} and consider the Wasserstein distance instead of f-divergence to avoid the requirement that $P$ is absolute continuous with respect to $\hat{P}_N$, which is impractical for continuous distribution.}:
\begin{align*}
d_\mathrm{W}(Q_1,Q_2):=\inf_\Pi\bc{\int_{\Xi\times \Xi}c(\xi_1,\xi_2)\Pi(\mathrm{d}\xi_1,\mathrm{d}\xi_2)}, 
\end{align*}
where $\Pi$ is a joint distribution over $(\xi_1,\xi_2)$, with its marginal distributions on $\xi_1$ and $\xi_2$ being $Q_1$ and $Q_2$ respectively; the cost function $c(\cdot,\cdot)$  used for  Wasserstein distance is chosen as the type-I version, with $c(x,y)=\|x-y\|_2$.

Let $\hat{x}_N$ be the solution derived from the RS model \eqref{eq:rs}, the reformulation of which will be elaborated upon in Section \ref{sec:tractability}. Our goal is to provide statistical  guarantees on $\hat{x}_N$ and $J^*$.

\subsection{Reference Value $\tau$}
\label{sec:reference_value}
The reference value $\tau$ introduced in the RS model \eqref{eq:rs}  is critical in controlling the robustness of the learned solution $\hat{x}_N$. Conceptually, following the satisficing criterion, the RS model ensures that any excess beyond the reference value $\tau$ under a certain distribution is controlled by a multiple of the distance between this distribution and the empirical distribution of the data. 
A larger $\tau$ indicates increased robustness considered in the RS model.

Choose $P$ as $\hat{P}_N$ in \eqref{eq:rs}, we easily obtain:
\begin{equation}\label{tau greater than optimal}
    \tau \geq \bbE_{\hat{P}_N}[h(x,\xi)].
\end{equation}
Inspired by this, \citeauthor{long2023robust} suggest choosing $\tau$  as:
\begin{align}\label{tau epsilon}
    \tau_{\epsilon}:= (1+\epsilon)\inf_x\bbE_{\hat{P}_N}[h(x,\xi)],
\end{align}
where $\epsilon$ is referred to as ``tolerance rate'' that the RS model allows for excess empirical loss.
This means that the reference value $\tau$, which we choose or tolerate, is $\epsilon$ more than the smallest cost achievable under the empirical distribution. 
We adopt this approach, focusing on characterizing the role of $\epsilon$  in the statistical guarantees provided by the RS model. Additionally,  $\epsilon$ will be the primary hyperparameter we adjust and analyze in the numerical experiments section.

\subsection{Reformulation}
\label{sec:tractability}
 The original RS optimization \eqref{eq:rs} requires enumerating over all possible distributions over $\mathbb{P}$, which may not be tractable. We now reformulate the  model \eqref{eq:rs}, following the practice by \citeauthor{long2023robust}. Let $\eta$ and $\xi$ be samples from $P$ and $\hat{P}_N$ respectively,  and let $\pi(\eta|\epsilon)$ be the conditional distribution of $\eta$ when conditioning on $\xi$. We have:
\begin{equation}
\label{eq:reformulation}
\begin{aligned}
    &\sup_P\lbrace \bbE_P[h(x,\eta)]-kd_W(P,\hat{P}_N)\rbrace\\
    =&\sup_{\pi}\iint[h(x,\eta)-kc\left(\xi,\eta\right)]d\hat{P}_N\left(\xi\right)d\pi\left(\eta|\xi\right)\\
    =&\bbE_{\hat{P}_N}[\sup_{z\in\Xi}h(x,z)-kc(\xi,z)],
\end{aligned}    
\end{equation}
where the last  equation  is achieved by choosing the maximizer $\pi$ as the Dirac distribution, which concentrates the mass at the point to maximize $\{h(x,\cdot)-kc(\xi,\cdot)\}$.
Then the RS model \eqref{eq:rs} can be reformulated as:
\begin{align}
\label{eq:rs_reformulated}
\min\quad & k\geq 0 \\
    \text{s.t.} &\quad \bbE_{\hat{P}_N}[\sup_{z\in\Xi}h(x,z)-kc(\xi,z)]\leq \tau. \nonumber \\
    &x \in \mathcal{X} \nonumber
\end{align}

With that, the optimizer $\hat{x}_N$ of RS model can be obtained in a hierarchical way. First, for a fixed decision variable $x$, let $k_{\tau}(x)$ be the smallest $k$ that satisfies the RS constraint:
\begin{align*}
\centering
& k_{\tau}(x):=\min k(x), \\ 
\text{s.t. } &
\bbE_{\hat{P}_N}[\sup_{z\in\Xi}h(x,z)-kc(\xi,z)]\leq \tau.
\end{align*}
Then $\hat{x}_N$ is the minimizer of  $k_\tau(x)$:
\begin{align}\label{mink}
    &\hat{x}_N:=\text{argmin}_x k_{\tau}(x).
\end{align}

We note that a similar reformulation technique has been employed by  \citeauthor{blanchet2019quantifying} to derive tractable solutions for DRO. However   in the DRO framework, the distribution is restricted to an ambiguity set, necessitating the use of a Lagrange multiplier for constraint conditions and the existence of strong duality. These constraints introduce additional assumptions, including the continuity of functions $h(x,\xi)$ and $c(\cdot,\cdot)$. In contrast, robust satisficing, which does not limit the distribution set, avoids these extra assumptions.

\section{Statistical Properties}
This section presents our main results for the statistical properties of the optimizer $\hat{x}_N$ in the RS model \eqref{eq:rs}.
We start by describing the  assumptions required for our analysis.

\begin{assumption}[Exponential tail decay in  random variable]\label{assump:light_tail}
There exists an $ a>1$,  such that $\bbE_{P^*}\bb{\exp(||\xi||^a)} <\infty$.
\end{assumption}
Assumption \ref{assump:light_tail} requires that $\xi$ is relatively light-tailed. It plays a key role in bounding the rate at which  the empirical distribution $\hat{P}_N$ approximates the true distribution $P^*$ under  the type-1 norm Wasserstein distance \cite{fournier2015rate} (see Proposition \ref{dw bound} for details). This assumption is relatively mild and is applicable to a broad range of distributions including sub-Gaussian random variables.

\begin{assumption}[Lipschitz continuity of loss function]\label{assump:lips} The loss 
    $h(x,\xi)$ is Lipschitz  with a uniform  constant $L$ in $\xi$.
\end{assumption}

Assumption \ref{assump:lips}  is essential for deriving the dual expression form of the type-1 norm Wasserstein distance  \cite{esfahani2015data} (see Proposition \ref{dual w-dis} for details). This assumption holds true for a wide range of machine learning problems, which we further elaborate in Appendix \ref{sec:ml_loss}. Note that we don't need the Lipschitz continuity assumption of $h(x,\xi)$
 with respect to $x$, which we defer the detailed discussion in the Appendix \ref{sec:lip continuity}.

\subsection{Confidence Intervals of Optimal Loss}
This section provides both non-asymptotic and asymptotic confidence intervals for the optimal loss $J^*$, the smallest attainable expected loss as defined in \eqref{J^*}, and the true loss of $\hat{x}_N$.

\begin{theorem}[Confidence intervals of optimal loss]\label{thm:confidence_intervals}
    
Suppose Assumptions \ref{assump:light_tail} \& \ref{assump:lips} hold. 
For any $N$, let $\beta_N$ be the confidence level. We have with probability at least $1-\beta_N$: 
\begin{equation}\label{thm p2}
    -L\cdot r_N +\frac{\tau_\epsilon}{1+\epsilon} \leq J^*\leq  \bbE_{P^*}[h(\hat{x}_N,\xi)]  \leq k_{\tau_\epsilon}\cdot r_N +\tau_{\epsilon},
\end{equation}

where 
$r_N$, denoted as the ``remainder'', is solved from the below equation: 
\begin{equation*}
    \beta_N = \left\{
    \begin{aligned}
        &c_1\exp\big(-c_2N{r_N}^{\max\{m,2\}}\big)&& \mbox{if }\:r_N\leq1,\\
        &c_1\exp\big(-c_2N{r^a_N}\big)&& \mbox{if } \:r_N>1, 
    \end{aligned}
    \right.
\end{equation*}
with $c_1, c_2$ as positive constants that only depend on exponential decay rate $a$ and dimension $m$.

Moreover, when choosing  the confidence sequence $\{\beta_N\}$ satisfying
$\sum_{N=1}^{\infty}\beta_N < \infty$, we have
\begin{align} \label{eq:thm_p4_general}
    P\Big\{ -L\cdot r_N +\frac{\tau_\epsilon}{1+\epsilon} \leq J^*\leq \bbE_{P^*}[h(\hat{x}_N,\xi)] \nonumber\\\leq k_{\tau_\epsilon}\cdot r_N +\tau_{\epsilon} \text{ for all sufficiently large }N\Big\} =1 .
\end{align}

\end{theorem}

Table \ref{tab:beta} outlines typical selections of $\beta_N$ and their respective rates of decay for the remainder $r_N$. Notably, the last two $\beta_N$ options satisfy $\sum_{N=1}^{\infty}\beta_N < \infty$ and $\lim_{N\rightarrow \infty}r_N=0$, under which  \eqref{eq:thm_p4_general} suggests asymptotic consistency of $\bbE_{P^*}[h(\hat{x}_N,\xi)]$(Note that this asymptotic interval applies to $J^*$ directly by  the convergence of empirical loss to the true loss as $N$
 increases):
\begin{align}\label{thm p4}
    P\Big\{\frac{\tau_\epsilon}{1+\epsilon} \leq  \bbE_{P^*}[h(\hat{x}_N,\xi)] \leq \tau_{\epsilon} \text{ for all} \nonumber\\\text{sufficiently large  }N\Big\} =1.
\end{align}
We recognize the challenge posed by the curse of dimensionality, as indicated by the exponent $m$ in $r_N$, which is a common issue associated with the Wasserstein distance \cite{esfahani2015data,kuhn2019wasserstein}, and we leave as a promising future research question.

We also note that the upper bound in Eq. \eqref{thm p2} includes $k_{\tau_\epsilon}$, which may be difficult to derive analytically.  Fortunately, the following lemma provides an upper bound guarantee for $k_{\tau_\epsilon}$.

\begin{lemma}[Fragility Upper Bound]\label{lemma:k tau leq L} Under Assumption \ref{assump:lips}, we have
$k_{\tau}\leq L$, where $k_\tau$ is solved from the RS model \eqref{eq:rs}.
\end{lemma}
Lemma \ref{lemma:k tau leq L} that we prove is  noteworthy on its own.
As pointed out by  \citeauthor{long2023robust}, $k_\tau$ characterizes the fragility of the model, with lower values indicating more robustness. Lemma \ref{lemma:k tau leq L} sets an upper bound for $k_\tau$ based on the Lipschitz constant $L$, suggesting the model fragility being controlled.

\begin{remark}\label{A_N}
In the proof detailed in the \cref{appendix proof of confidence intervals}, we  establish the following relationship:
\begin{equation}\label{proof chain}
\begin{split}
    -L\cdot d_W(P^*,\hat{P}_N)+\frac{\tau_\epsilon}{1+\epsilon} \leq J^* \leq \bbE_{P^*}[h(\hat{x}_N,\xi)]\\\leq k_{\tau_\epsilon}\cdot d_W(P^*,\hat{P}_N)+\tau_{\epsilon}.
\end{split}
\end{equation}   

Equation~\eqref{proof chain} illustrates that 
the true loss of $\hat{x}_N$ (the optimizer  obtained from the RS model), under the target distribution $P^*$, also falls within
 the confidence interval provided by Equation \eqref{thm p2}. Furthermore, Equation \eqref{thm p2} further facilitates the derivation of the upper bound of the generalization error in Theorem \ref{thm:generalization_error}.
    
\end{remark}
\begin{remark}

Equation \eqref{thm p2} provides a guideline on determining the sufficient sample size required to achieve a predefined accuracy at a specified confidence level $\beta_N$. This sample size  is primarily quantified by the width of the confidence interval and mainly driven by $r_N$. Table \ref{tab:beta} illustrates various selections of $\beta_N$ along with their corresponding $r_N$ values, which allows us to explicitly compute the sample size required for specific scenarios.
\end{remark}

 By integrating Lemma \ref{lemma:k tau leq L} with Theorem \ref{thm:confidence_intervals}, we derive a simpler  form of confidence intervals for $J^*$, which depends solely on the Lipschitz constant $L$ and the reference value $\tau_\epsilon$,   eliminating the need to compute $k_{\tau_\epsilon}$ from the RS model.
\begin{corollary}
\label{cor:loose_interval}
    Suppose Assumptions \ref{assump:light_tail} \& \ref{assump:lips} hold. For any $N$ and the confidence level $\beta_N$, let $r_N$ be solved as in Theorem \ref{thm:confidence_intervals}. With probability at least $1-\beta_N$, we have
    \begin{align*}
    -L\cdot r_N +\frac{\tau_\epsilon}{1+\epsilon} \leq J^*, \bbE_{P^*}[h(\hat{x}_N,\xi)] \leq L\cdot r_N +\tau_{\epsilon}.
    \end{align*}
\end{corollary}
The remainder  $r_N$ becomes negligible for  choices of $\beta_N$ listed in Table \ref{tab:beta}. Thus Corollary \ref{cor:loose_interval} indicates that as $N$ approaches $\infty$, the expected loss $E_{P^*}h(\hat{x}_N,\xi)$ of the optimizer $\hat{x}_N$ will also fall within the interval $[\frac{\tau_\epsilon}{1+\epsilon}, \tau_\epsilon]$. This allows us to characterize the loss value that the optimizer can achieve and also shows that the regret of our optimizer $\hat{x}_N$ (the gap between $E_{P^*}h(\hat{x}_N,\xi)$ and the true loss) will be controlled by the length of the interval asymptotically.

\begin{table*}
    \centering
    \caption{Choices of Confidence Level $\beta_N$}
    \begin{tabular}{c|c}
    \hline

    Choice of $\beta_N$& Corresponding $r_N$ \\ \hline
          $\beta_N\equiv\beta$  &${r_N= \left\{\begin{array}{ll}\Big(\frac{\log(c_1\beta^{-1})}{c_2N}\Big)^{1/\max\{m,2\}}
&if\:N\geq\frac{\log(c_1\beta^{-1})}{c_2},\\\Big(\frac{\log(c_1\beta^{-1})}{c_2N}\Big)^{1/a}&if \:N<\frac{\log(c_1\beta^{-1})}{c_2}. 
\end{array}\right.}$ \\ \hline
       $\beta_N=\exp(-\gamma\sqrt{N})$, $\gamma>0$ & $r_N= \left\{\begin{array}{ll}\Big(\frac{\log c_1}{c_2N}+\frac{\gamma}{c_2\sqrt{N}}\Big)^{1/\max\{m,2\}}
&if\:c_2N-\gamma\sqrt{N}\geq\log c_1,\\\Big(\frac{\log c_1}{c_2N}+\frac{\gamma}{c_2\sqrt{N}}\Big)^{1/a}&if\:c_2N-\gamma\sqrt{N}<\log c_1.
\end{array}\right.$ \\ \hline
   $\beta_N=N^{-\alpha}$, $\alpha>0$ & $r_N=\left\{\begin{array}{ll}\Big(\frac{\log c_1}{c_2N}+\alpha\frac{\log N}{c_2N}\Big)^{1/\max\{m,2\}}
&if\:c_2N-\alpha\log{N}\geq\log c_1,\\\Big(\frac{\log c_1}{c_2N}+\alpha\frac{\log N}{c_2N}\Big)^{1/a}&if\:c_2N-\alpha\log{N}<\log c_1. 
\end{array}\right.$  \\ 
    \bottomrule
    \end{tabular}
    
    \label{tab:beta}
\end{table*}

To conclude this section, we offer a brief comparison of our confidence intervals with those derived by \citeauthor{esfahani2015data}. In their DRO framework,  they define
\begin{equation*}
  \tilde{J}_N=\inf_{x\in \mathcal{X}}\sup_{P\in B(\hat{P}_N,\epsilon(\beta_N))} \bbE_P\bb{ h(x,\xi)},
\end{equation*}
where $B(\hat{P}_N,\epsilon(\beta_N))$ represents a Wasserstein ball with its center $\hat{P}_N$ and radius $\epsilon(\beta_N)$.
Under similar assumptions, \citeauthor{esfahani2015data} show that
\begin{align*}
 P\{J^*\leq \tilde{J}_N\}\geq 1-\beta_N,
\end{align*}
which provides only an upper bound for the optimal loss $J^*$. Moreover, this upper bound $\tilde{J}_N$ requires to solve the minimax problem in the DRO framework. 
In contrast, our confidence intervals from Corollary \ref{cor:loose_interval} are derived through the relatively easier optimization of the ERM problem than the minimax problem, and our results provide two-sided rather than one-sided confidence intervals.

\subsection{Finite-Sample Generalization Error Bound}
We now focus on characterizing the generalization error of the optimizer $\hat{x}_N$ derived from the RS model. The generalization error, denoted as $R(P^*, \hat{x}_N)$, is defined as follows:
\begin{align*}
    R(P^*, \hat{x}_N):=&\bbE_{P^*}[h(\hat{x}_N,\xi)]-J^*\\
    =&\bbE_{P^*}[h(\hat{x}_N,\xi)]-\bbE_{P^*}[h(x^*,\xi)].
\end{align*}

\begin{theorem}\label{thm:generalization_error}
 Suppose Assumptions \ref{assump:light_tail} \& \ref{assump:lips} hold. With probability at least $1-\beta_N$, we have:
\begin{align}
    R(P^*, \hat{x}_N) \leq \epsilon\cdot J^*+(2+\epsilon)\cdot L\cdot r_N,
\end{align}
where $r_N$ is the reminder  solved as in Theorem \ref{thm:confidence_intervals}.

Taking  expectation with respect to data, we have:
\begin{align}
    \bbE_{P^*}\bb{R(P^*, \hat{x}_N)} \leq \epsilon\cdot J^*+O(L\cdot N^{-\min\{\frac{1}{m}, \frac{1}{2} \}}).
\end{align}

\end{theorem}

\begin{remark}
We further elaborate on the ``expectation with respect to data".
Recall that we derive the optimizer $\hat{x}_N$ based on sample data, which are random variables that follow the source distribution $P^*$. As a result, the $\hat{x}_N$ and its generalization error upper bound are also random variables. So we take the expectation with respect to the randomness from the sample data to derive our expected version of the generalization error upper bound. 

\end{remark}

Theorem \ref{thm:generalization_error} explicitly characterizes  how the generalization error is influenced by $\epsilon$. By reducing $\epsilon$ as the sample size $N$ increases — indicating less tolerance for empirical loss excess with more data — we can  bound the generalization error more succinctly, as outlined in the following result.

\begin{corollary}\label{cor:generalization_error}
 Suppose Assumptions \ref{assump:light_tail} \& \ref{assump:lips} hold. Choose reference value $\tau_{\epsilon_N}$ with $\epsilon_N=N^{-\min\{\frac{1}{m}, \frac{1}{2} \}}$. Then 
\begin{align}
    \bbE_{P^*}\bb{R(P^*, \hat{x}_N)} = O(L\cdot N^{-\min\{\frac{1}{m}, \frac{1}{2} \}}).
\end{align}
\end{corollary}

\section{Guarantees under Distribution Shift}
As discussed, the distribution selection under the RS framework is globalized, eliminating the need to pre-select a radius to restrict the distribution domain. We take this advantage further and integrate it into the earlier derivation process, allowing us to straightforwardly derive the confidence intervals and the finite-sample generalization error bound under distribution shifts.

Consider that  samples are drawn from the source distribution $P^*$, and the empirical distribution is denoted as $\hat{P}_N$. The decision variable learned from the RS model  \eqref{eq:rs} is $\hat{x}_N$. Under distribution shifts, we  evaluate the performance when applying $\hat{x}_N$ to another distribution $\tilde{P}$, which   may shift from $P^*$, resulting in a certain degree of discrepancy.

Define the optimal loss  under the new distribution $\tilde{P}$ as $\tilde{J}$:
\begin{align*}
    \Tilde{J}:=\inf_{x\in\mathcal{X}}\bbE_{\Tilde{P}}[h(x,\xi)]=\bbE_{\tilde{P}}[h(\Tilde{x},\xi)],
\end{align*}
where $\Tilde{x} = \text{argmin}_{x}\bbE_{\Tilde{P}}[h(x,\xi)]$. For the learned decision variable $\hat{x}_N$, denote the corresponding generalization error as 
\begin{align*}
    R(\Tilde{P}, \hat{x}_N):=~&\bbE_{\Tilde{P}}[h(\hat{x}_N,\xi)]-\tilde{J}\\
    =~&\bbE_{\Tilde{P}}[h(\hat{x}_N,\xi)]-\bbE_{\Tilde{P}}[h(\Tilde{x},\xi)].
\end{align*}
Our goal is to derive confidence intervals for $\tilde{J}$ and generalization error bound for  $R(\Tilde{P}, \hat{x}_N)$.

\begin{theorem}[Distribution Shift]\label{dis. shift}
Suppose Assumptions \ref{assump:light_tail} \& \ref{assump:lips} hold. 
For any $N$, let $\beta_N$ be some nominal confidence level. We have with probability at least $1-\beta_N$: 
\begin{align*}
    -L\cdot r_N- L \cdot d_W(P^*,\Tilde{P}) +\frac{\tau_\epsilon}{1+\epsilon} \leq \Tilde{J} \leq \bbE_{\Tilde{P}}h(\hat{x}_N,\xi)\\ \leq k_{\tau_\epsilon}\cdot r_N + k_\tau \cdot d_W(P^*,\Tilde{P})+\tau_{\epsilon},
\end{align*}
and
\begin{align*}
    R(\Tilde{P}, \hat{x}_N) \leq \epsilon\cdot \Tilde{J}+(2+\epsilon)\cdot L\cdot d_W(P^*,\Tilde{P})\\
    +(2+\epsilon)\cdot L\cdot r_N,
\end{align*}
where the reminder $r_N$ is solved as Theorem \ref{thm:confidence_intervals}.

Taking the expectation on data, we have:
\begin{align*}
    \bbE_{{P^*}}\bb{R(\Tilde{P}, \hat{x}_N)} \leq \epsilon\cdot \Tilde{J}+ (2+\epsilon)\cdot L\cdot d_W(P^*,\Tilde{P})\\
    +O\bp{L\cdot N^{-\min\{\frac{1}{m}, \frac{1}{2} \}}}.
\end{align*}
\end{theorem}
This theorem shows that results under distribution shifts  merely require adding a  multiple of the  shift distance.

\begin{remark}
   While our results face the common curse of dimensionality issue associated with the Wasserstein distance, they embody a trade-off. In higher dimensions, despite the slow decay of the remainder term $r_N$, a greater degree of distribution shift is tolerable. Specifically, when the distribution shift decays at the rate of $N^{-\min\{\frac{1}{m}, \frac{1}{2}\}}$, this rate can be integrated with the remainder term to yield the following guarantee:
\begin{align*}
    \bbE_{{P^*}}\bb{R(\Tilde{P}, \hat{x}_N)} \leq \epsilon\cdot \Tilde{J}+O(L \cdot N^{-\min\{\frac{1}{m}, \frac{1}{2} \}}).
\end{align*}
This implies that the RS model can accommodate a distribution shift up to $N^{-\min\{\frac{1}{m}, \frac{1}{2}\}}$ while still maintaining performance comparable to scenarios with no shift.
\end{remark}

Finally, we compare our results with DRO. 
Under the DRO framework, if the distribution shifts, we must require the radius to reach a certain magnitude so that the ambiguity set can contain the distribution after the shift.
However, as this ball expands, the worst-case expected value within DRO's conservative minimax framework deteriorates.   
In contrast, the RS framework, benefiting from its globalized distribution selection, only requires the inclusion of a linear multiple of the shift distance to address the same situation.

\section{Numerical Experiments}
\label{sec:numerical}
In this section, we conduct numerical evaluations of the RS model under both the original sampling distribution and distributional shifts. We compare RS with the baseline method, which is empirical risk minimization (ERM). Additionally, we establish connections with DRO and demonstrate that RS exhibits lower sensitivity to hyperparameter tuning.

All experiments are based on a data generating process detailed below. We define  the random variable $\xi$ as  $\xi=(u,y)$, with $u\in \bbR^{m_u}$ representing the feature variable and $y\in\bbR$ as the label variable.  The sampling distribution $P^*$ is specified as follows: the feature variable $u$ is drawn from a normal distribution:
\begin{equation}
\label{eq:sample_independent_var}
    u\sim \mathcal{N}\bp{[0.5,0.5,...,0.5]^T,0.5I_{m_u} };
\end{equation}
and the label variable $y$ is generated via a linear model:
\begin{equation*}
    y = u\cdot x^* +e,
\end{equation*}
where   $\cdot$ means the inner product, $x^*$ is the true model parameter, and $e$ is the exogenous noise sampled from $\calN(0,0.1)$.  
$P^*$ satisfies Assumption \ref{assump:light_tail} because Gaussian distribution is light-tailed.
Let the training data $\bc{(u_i, y_i)}_{i=1}^N$ be i.i.d. samples from the  distribution $P^*$. 

We use  $\ell_1$ loss for model parameter $x$: $h(u,y,x)=|y-u\cdot x|$, which satisfies the Lipschitz condition in Assumption \ref{assump:lips}. For the cost function used in the type-I Wasserstein distribution, we follow \cite{blanchet2019robust} and slightly modify  its original definition of the $l_2$ norm as follows\footnote{The purpose of this adjustment is to make the subsequent exposition more concise. We leave the results under the original $l_2$ norm in \cref{appendix:Other Equivalent Conclusions}}:
$$c(\xi_1,\xi_2)=c((u_1,y_1),(u_2,y_2))=\Big\{\begin{array}{ll}
      ||u_1-u_2||_2 & \mbox{if }y_1=y_2,\\
      +\infty  & \mbox{otherwise}.
\end{array}$$

The learned parameter $\hat{x}_N$ is evaluated on  the target distribution $\tilde{P}$. The marginal ditribution of $u$ under $\tilde{P}$ is identical to that under $P^*$, following \eqref{eq:sample_independent_var}. However, the label variable $y$ is generated under a potentially different parameter $\tilde{x}$:
\begin{equation*}
    y = u\cdot\tilde{x} +e.
\end{equation*}
In the following sections, we will evaluate the performances of RS  under two scenarios: when $P^*=\tilde{P}$ (i.e., $x^*=\tilde{x}$), representing settings without distribution shift, and when $P^*\neq \tilde{P}$ (i.e., $x^*\neq\tilde{x}$), indicating settings with distribution shift. We focus on the mean square error (MSE) in the target distribution as the performance metric. We will conclude this section by drawing connections between  RS and DRO.

\subsection{RS Performance in the Sampling Distribution}\label{sec: RS performance on sample}
In this section, we evaluate the RS optimizer $\hat{x}_N$  under the sampling distribution.
Although the target distribution aligns with the sampling distribution,
discrepancies between the empirical and sampling distributions may arise, particularly in small-sample regimes for high dimensional random variables. For this purpose, we consider a relatively high-dimensional setting with the dimension $m_u=10$ and the true model parameter $\tilde{x}=x^*=[2.0, -1.0,...,2.0,-1.0]^T$. 
We  investigate the generalization performance of  $\hat{x}_N$ across various sample sizes. 

\begin{figure}
    \centering
    \includegraphics[scale = 0.4]{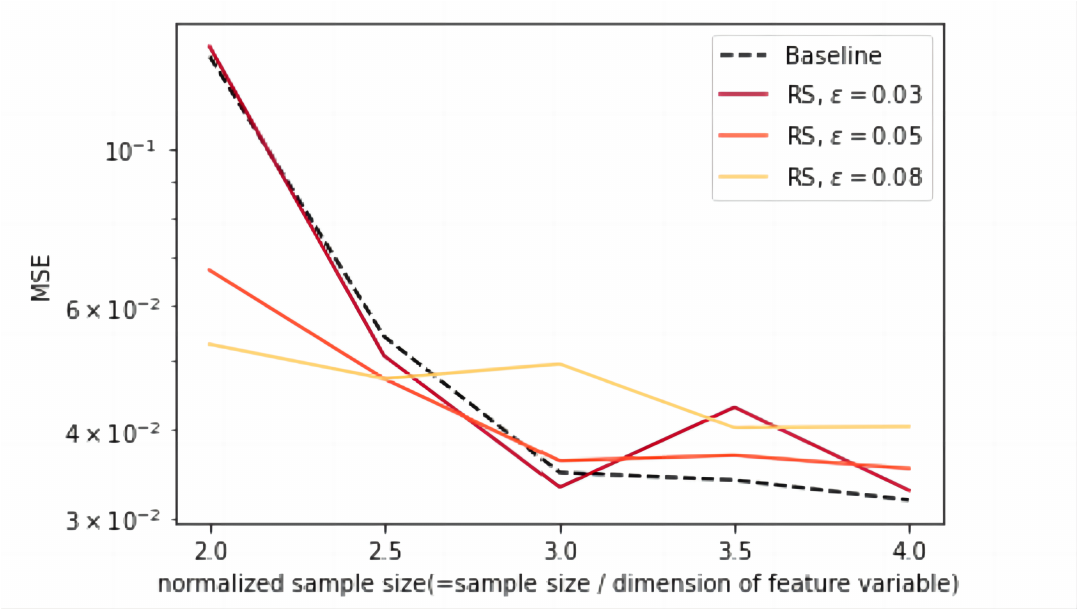}
    \caption{Performances across various sample sizes. RS outperforms the ERM baseline in small-sample regimes.}
    \label{fig:RS_sample_size}
\end{figure}

\Cref{fig:RS_sample_size} 
demonstrates how the RS model's performance varies with different settings of the tolerate rate $\epsilon$ and across  various  sample sizes.
For smaller sample sizes, 
the RS model outperforms the baseline that does not incorporate robustness; among those RS models, those configured with a larger $\epsilon$ (indicating  a greater emphasis on robustness) perform better. 
This result notes the importance of accounting for robustness, particularly when there is a notable gap between the sampling and empirical distributions in  small-sample regimes.
As the sample size increases, the relative benefit of the RS model decreases. 
This trend is expected since a larger dataset allows to better picture the sampling distribution, making the baseline approach of empirical risk minimization increasingly effective.

\subsection{RS Performance under Distribution Shift}

\label{sec:numerical_shift}

We now evaluate the performance of RS under distribution shift. Let the model parameter in the sampling distribution be $x^*=[2.0,-1.0]$, and the  model parameter in the  target distribution  be:
\[
\tilde{x}=[2.00-0.05\times \textsc{Degree}, -1.00+0.025\times \textsc{Degree}],\]
where \textsc{Degree} is a positive number characterizing the degree of distribution shift\footnote{Here our focus is on evaluating robustness in scenarios  with smoothed parameters, rather than under arbitrary perturbations of $x^*$. This setup is based on  the observation that  both DRO and RS, known for their robustness,  typically yield smoother parameters than those derived from   direct empirical risk minimization.
}. 
As \textsc{Degree} increases, the discrepancy between $x^*$ and $\tilde{x}$ increases, leading to a larger distribution shift between the sampling distribution $P^*$ and the target distribution $\tilde{P}$.

\begin{figure}
    \centering
    \includegraphics[scale = 0.4]{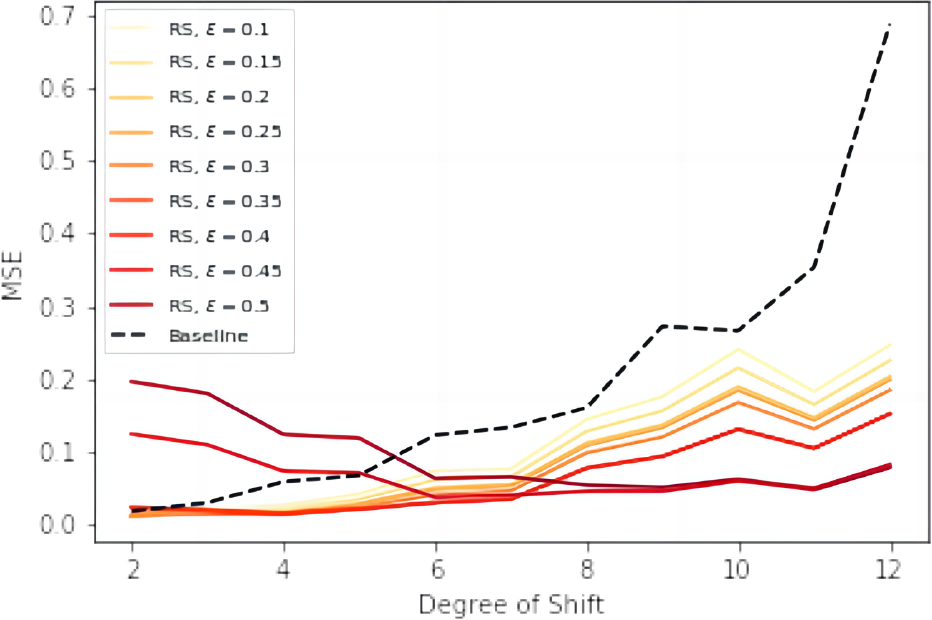}
    \caption{Performances across various degree of  distribution shifts. RS outperforms the ERM baseline under distribution shifts.}
    \label{fig:RS_shift}
\end{figure}

\cref{fig:RS_shift} shows how the RS model performs when configured with different  tolerance rates $\epsilon$ and under various distribution shifts. For minor distribution shifts (\textsc{Degree} less than $2$), the RS model's performance is comparable to the baseline, deteriorating slightly at models of larger $\epsilon$ for stronger robustness control. However, with more substantial distribution shifts (\textsc{Degree} greater than $5$), the RS model almost consistently outperforms the baseline, presenting stronger robustness under larger distribution shifts. This result  highlights the   potential of RS framework for strong generalization guarantee in uncertain environments.

\subsection{Connection to DRO under  Lipschitz loss}

We proceed to compare RS and DRO. We establish explicit correspondence between the hyperparameters of RS and DRO in this type of problem. And we conduct experiments to compare their  sensitivities to hyperparameter tuning.

\subsubsection{Hyperparameter correspondence}

Consider a Lipschitz loss function $L(\cdot)$. We reformulate the general  Lipschitz-loss learning problem following the  DRO literature \cite{blanchet2019robust,shafieezadeh2019regularization}:
\begin{align}
\label{eq:dro_equivalent}
\min_x    \frac{1}{N}\sum_{i=1}^{N}L(y_i-u_i\cdot x)+ r\cdot ||x||_2.
\end{align}
Building on the reformulation presented in Section \ref{sec:tractability}, this Lipschitz-loss learning problem under the RS framework is equivalent to:
\begin{align}
\label{RS equivalent}
\min_x ~&||x||_2, \\
\mbox{s.t.}~ &\frac{1}{N}\sum_{ i=1}^{N}L(y_i-u_i\cdot x)\leq \tau. \nonumber
\end{align}
By applying the Lagrangian method to solve Equation \eqref{RS equivalent} and with the strong duality held, we further deduce that Equation \eqref{RS equivalent} simplifies to the following expression (see \cref{appendix derivation of equaivalent models} for the proof):
\begin{align}
\label{eq:RS_equivalent_2}
\sup_{\lambda>0}\inf_{x}\frac{\frac{1}{N}\sum_{ i=1}^{N}L(y_i-u_i\cdot x)+\lambda\cdot||x||_2-\tau}{\lambda}.
\end{align}
We immediately observe a clear link between RS and DRO for the general Lipschitz-loss learning problem: 
given a reference value $\tau$, solving the RS model  \eqref{eq:RS_equivalent_2} yields the optimizer $(\hat{\lambda}_N, \hat{x}_N)$.
Then, by   setting the radius $r$ in the DRO model \eqref{eq:dro_equivalent} to $\hat{\lambda}_N$, the DRO model \eqref{eq:dro_equivalent} generates  the same optimizer for the model parameter $x$.
Recall that the reference value is set to be $\tau_\epsilon = (1+\epsilon)\inf_x\bbE_{\hat{P}_N}[h(x,\xi)]$ throughout this paper, where $\epsilon$ is the tolerate rate  that controls the robustness of RS. 
Thus  each hyperparameter  $\epsilon$ in the RS model is associated with a specific radius $r(\epsilon)$, the hyperparameter in the DRO model.

\begin{figure}
    \centering
    \includegraphics[scale = 0.03]{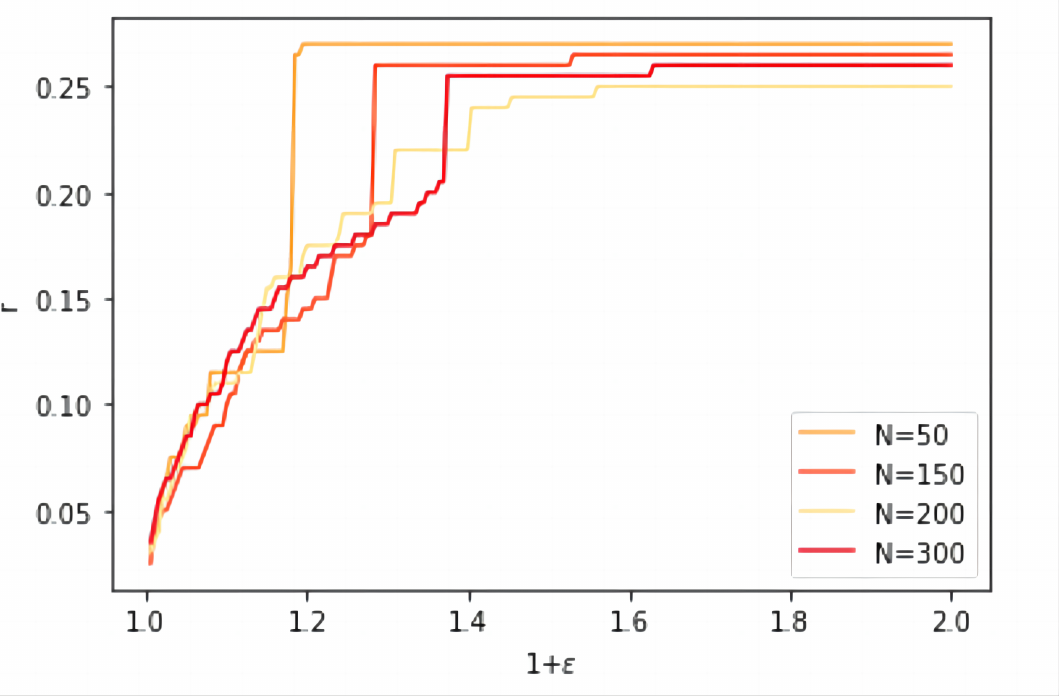}
    \caption{Correspondence between RS torelance rate parameter $\epsilon$ and DRO radius parameter $r$.}
    \label{r-e_function}
\end{figure}

\cref{r-e_function} illustrates the relationship between the robustness-controlling hyperparameters of the two models: the DRO radius $r$ and the RS torelance rate $\epsilon$.
Notably, the function $r(\epsilon)$ is concave with respect to $\epsilon$, flatting as   $1+\epsilon$ nears $1.6$.
This indicates that, to achieve comparable performance, the RS model can accommodate larger variations in $\epsilon$ compared to variations in $r$ for the DRO model, implying that RS is less sensitive to hyperparameter tuning.  This observation will be further supported by the following experiments.

\subsubsection{Numerical sensitivity analysis}
We now conduct experiments to evaluate the sensitivity of  RS and DRO  to  hyperparameter tuning. Specifically, we vary the tolerance rate $\epsilon$ in the RS model and the radius $r$ in the DRO model.
We set the model parameter in the sampling distribution to $x^*=[2.0, -1.0]^T$, as in Section \ref{sec:numerical_shift}; and set the target environment to be $\tilde{x}=[1.80,-0.90]^T$. 

\begin{figure}
    \centering
\begin{subfigure}
    \centering
    \includegraphics[scale = 0.4]{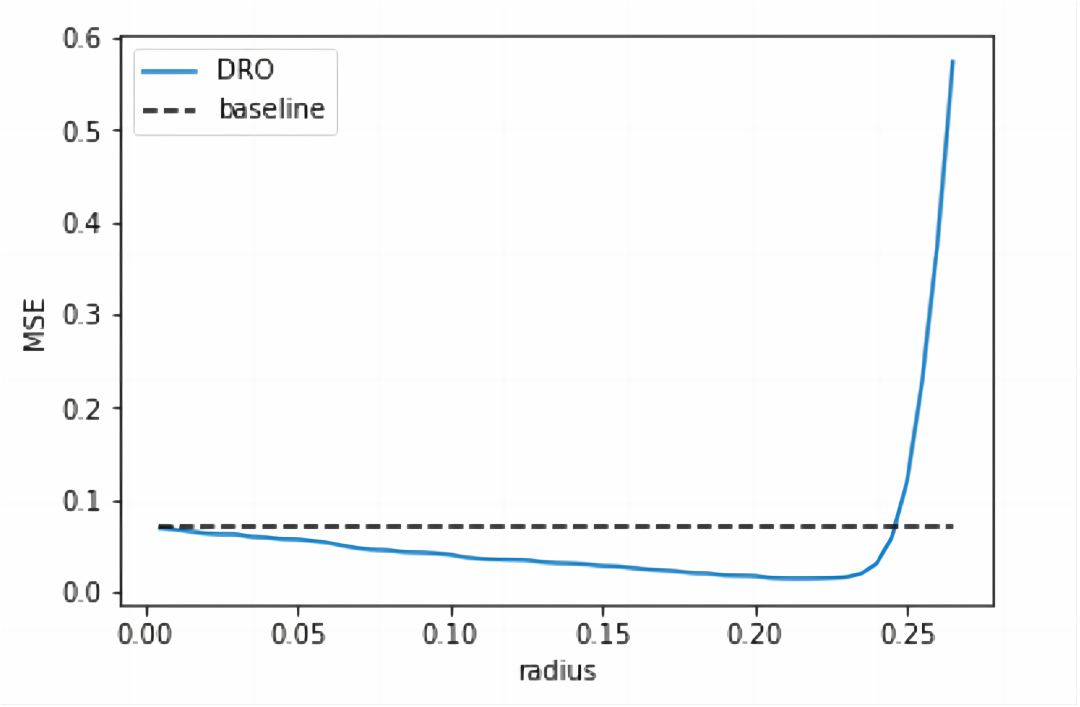}
    \caption{DRO performance under distribution shifts. DRO model shows  higher sensitivity to haperparameter $r$.}
    \label{DRO performance}
\end{subfigure}

\hfill

\begin{subfigure}
    \centering
    \includegraphics[scale = 0.4]{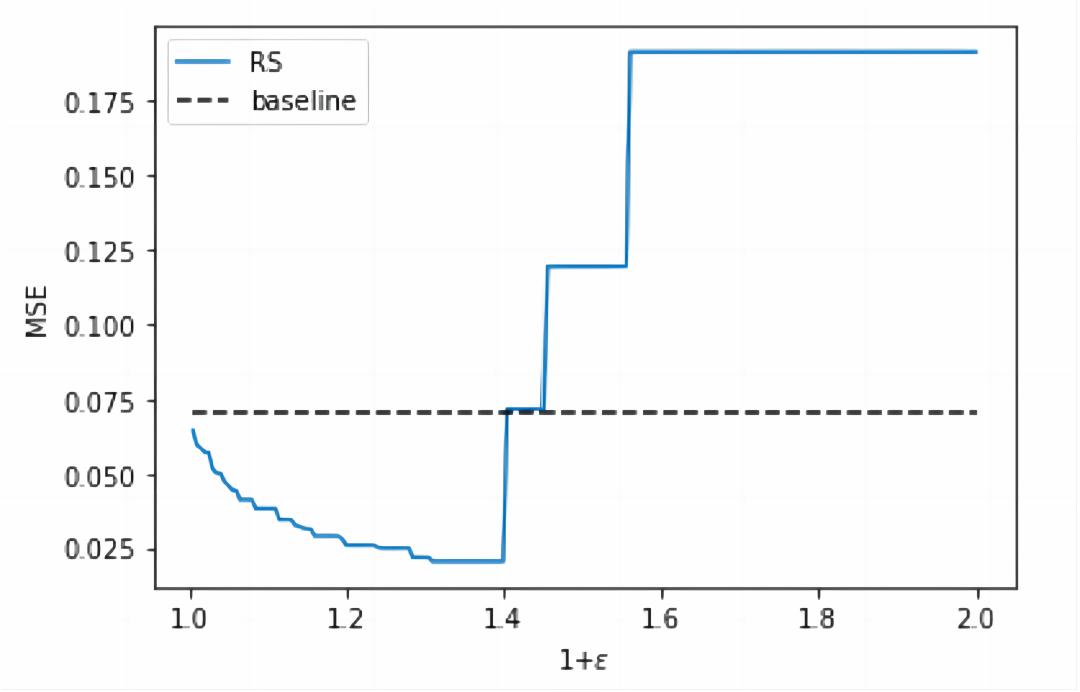}
    \caption{RS performance under distribution shifts. RS model shows lower sensitivity to hyperparameter $\epsilon$. }
    \label{RS performance}
\end{subfigure}

\end{figure}

\cref{DRO performance}
shows that DRO outperforms the baseline for radius smaller than $0.24$, with the optimal $r$ around $0.215$. 
\cref{RS performance}  shows that RS exceeds the baseline when $1+\epsilon$ is below $1.4$, with the optimal $1+\epsilon$ around $ 1.285$. 
The smallest MSEs from both DRO and RS models are comparable.

However, there is a drastic MSE surge in response to changes of $r$ for DRO in \cref{DRO performance}, in contrast to the milder variation of MSE to $\epsilon$ for RS showed in \cref{RS performance}. This difference in hyperparameter sensitivity aligns with  the nuanced relationship between $r$ and $\epsilon$ depicted in \cref{r-e_function}.
In particular, within a 15\% relative error range around the optimal hyperparameters, the MSE for DRO may spike to $0.30$, whereas RS maintains a more stable MSE of $0.125$. This suggests that RS offers greater  flexibility in setting the tolerate rate hyperparameter $\epsilon$, unlike DRO, which requires more precise tuning for the radius $r$.

\section{Conclusions}
This paper focuses on exploring the statistical properties of the RS model, a recent robust optimization framework introduced in \cite{long2023robust}. 
We provide theoretical guarantees for the RS model, including two-sided confidence intervals for the optimal loss and finite-sample generalization error bounds. 
These guarantees extend to scenarios involving distribution shifts, highlighting the RS model's robust generalization performance.
Our numerical experiments reveal the superiority of the RS model compared to the baseline empirical risk minimization method, particularly in small-sample regimes and under distribution shifts. 
We establish explicit connections between the RS and DRO frameworks within specific models, showcasing that  RS exhibits lower sensitivity to hyperparameter tuning than DRO, making it a more practical and interpretable choice.
Future research directions include extending our analysis of RS  to other distribution distances such as $f$-divergence, proving lower bounds for generalization error, and applying RS  to various practical applications.

\section*{Acknowledgement}
We would like to thank the reviewers for their constructive feedback, which has greatly improved the current version of our work. Yunbei Xu acknowledges support from ARO through award W911NF-21-1-0328 and from the Simons Foundation and NSF through award DMS-2031883. Ruohan Zhan is partly supported by the Guangdong Provincial Key Laboratory of Mathematical Foundations for Artificial Intelligence (2023B1212010001) and the Project of Hetao Shenzhen-Hong Kong Science and Technology Innovation Cooperation Zone (HZQB-KCZYB-2020083).

\section*{Impact Statement}
This paper presents work whose goal is to advance the field of Machine Learning. There are many potential societal consequences of our work, none which we feel must be specifically highlighted here.

\bibliographystyle{icml2024}

\newpage
\appendix
\onecolumn

\section{Key Propositions}

\begin{proposition}\label{dual w-dis}
For any distributions $\mathbb{Q}_1$, $\mathbb{Q}_2\in\mathcal{M}(\Xi)$, we have
\begin{align}\label{repre thm}  d_\mathrm{W}\big(\mathbb{Q}_1,\mathbb{Q}_2\big)=\sup_{f\in\mathcal{L}}\Big\{\:\int_{\Xi}f(\xi)\:\mathbb{Q}_1(\mathrm{d}\xi)-\int_{\Xi}f(\xi)\:\mathbb{Q}_2(\mathrm{d}\xi)\Big\}, 
\end{align}
where $\mathcal{L}$ denotes the space of all Lipschitz functions with $\begin{aligned}|f(\xi)-f(\xi')|\le\|\xi-\xi'\|\:for\:all\:\xi,\xi'\in\Xi.\end{aligned}$ and $||\cdot||$ is a norm 
\end{proposition}
This is the dual representation of Wasserstein distance, and it needs the Assumption \ref{assump:lips}.

We will also utilize the bound of Wasserstein distance between $P^*$ and $\hat{P}_N$, which is presented below.

\begin{proposition}\cite{fournier2015rate}\label{dw bound}
    If Assumption\ref{assump:light_tail} holds, we have

\begin{align*}
P^N\Big\{d_W\big(P^*,\hat{P}_N \big)\geq r\Big\}
\leq\left\{\begin{array}{ll}c_1\exp\big(-c_2Nr^{\max\{m,2\}}\big)&if\:r\leq1,\\c_1\exp\big(-c_2Nr^a\big)&if\:r>1,\end{array}\right.
\end{align*}

for all $N\geq1$, the dimension of $\xi$ : $m\neq2$ and $r>0$, where $c_{1},c_{2}$ are positive constants that only depend on $a$ and $m$.
\end{proposition}

\section{Proof of Lemmas and Theorems}
\subsection{Proof of Lemma \ref{lemma:k tau leq L}. }
Choose $x=\hat{x}_N$, we have 
\begin{align}\label{k tau leq L1}
    \bbE_{P}[h(\hat{x}_N, \xi)]-\tau \leq k_{\tau} d_W(P,\hat{P}_N) \quad \forall P \in \mathbb{P}.
\end{align}
Moreover, by the definition of $k_{\tau}$, for any $\delta>0$, we can choose one distribution $P_1$, which satisfies:
\begin{align}\label{k tau leq L2}
    \bbE_{P_1}[h(\hat{x}_N, \xi)]-\tau \geq (k_{\tau}-\delta)  d_W(P_1,\hat{P}_N).
\end{align}
Then \eqref{k tau leq L2} minus \eqref{k tau leq L1}, we have:
\begin{align}\label{k tau leq L3}
    (k_{\tau}-\delta) d_W(P_1,\hat{P}_N)-&k_{\tau} d_W(P,\hat{P}_N)\\
    &\leq \bbE_{P_1}[h(\hat{x}_N, \xi)]-\bbE_{P}[h(\hat{x}_N, \xi)]\\
    &\leq L\cdot d_W(P_1,P),
\end{align}
for all $\delta>0$ and $P\in\mathbb{P}$, Where the second inequality utilizes \eqref{repre thm}. Due to the arbitrariness of $\delta$, we can ignore the terms of $\delta$ and choose $\hat{P}_N$ as $P$ in \eqref{k tau leq L3}, we get:
\begin{align}\label{k tau leq L4}
    k_{\tau} d_W(P_1,\hat{P}_N)\leq L\cdot d_W(P_1,\hat{P}_N).
\end{align}
If $d_W(P_1,\hat{P}_N)>0$, we will complete the proof.\par

Now we explain why $d_W(P_1,\hat{P}_N)>0$ holds. Actually, $d_W(P_1,\hat{P}_N)=0$ if and only if $P_1=\hat{P}_N$, a.s. But if $P_1=\hat{P}_N$, then \eqref{k tau leq L2} leads to that $ \bbE_{\hat{P}_N}[h(\hat{x}_N, \xi)]\geq\tau$. Meanwhile, we can also choose $P=\hat{P}_N$ in \eqref{k tau leq L1} and we find $ \bbE_{\hat{P}_N}[h(\hat{x}_N, \xi)]\leq\tau$. So $ \bbE_{\hat{P}_N}[h(\hat{x}_N, \xi)]=\tau$. But it is impossible to hold because we can change $\epsilon$ in $\tau=\tau_\epsilon$ randomly.
$\hfill\blacksquare$

\subsection{Proof of Theorem \ref{thm:confidence_intervals}. } 
\label{appendix proof of confidence intervals}
For the right side , $J^*=\bbE_{P^*}[h(x^*,\xi)]\leq \bbE_{P^*}[h(\hat{x}_N,\xi)]\leq k_{\tau_\epsilon}\cdot d_W(P^*,\hat{P}_N)+\tau_{\epsilon}$, where the second inequality is derived from the model \eqref{eq:rs} itself and the fact that $k_{\tau}(\hat{x}_N)=k_{\tau}$. \\
For the other side, by the definition of $J^*$, for any $\eta>0$, choose $x_{\eta}$ satisfies:
\begin{align}\label{pf p1 1}
    \bbE_{P^*}[h(x^{\eta},\xi)] \leq J^*+\eta.
\end{align}
Then
\begin{align}
    \tau_\epsilon&=(1+\epsilon)\inf_x\bbE_{\hat{P}_N}[h(x,\xi)]\\&\leq(1+\epsilon)\bbE_{\hat{P}_N}[h(x_{\eta},\xi)]\\
    &\leq(1+\epsilon)\Big[L\cdot d_W(P^*,\hat{P}_N)+\bbE_{P^*}[h(x_{\eta},\xi)]\Big]\\
    &\leq (1+\epsilon)L\cdot d_W(P^*,\hat{P}_N)+(1+\epsilon)(J^*+\eta),
\end{align}
for all $\eta>0$. The second inequality is derived from  \eqref{repre thm} and the third inequality uses \eqref{pf p1 1}. Due to the arbitrariness of $\eta$, we have $\tau_\epsilon\leq(1+\epsilon)L\cdot d_W(P^*,\hat{P}_N)+(1+\epsilon)J^*$, which can be solved:\\
\[ -L\cdot d_W(P^*,\hat{P}_N)+\frac{\tau_\epsilon}{1+\epsilon} \leq J^*.\]

For \eqref{thm p2}, we can simply utilize  \eqref{dw bound} :
\begin{align}\label{d to p}
    P\Big\{-L\cdot r_N +\frac{\tau_\epsilon}{1+\epsilon} \leq J^* \leq k_{\tau_\epsilon}\cdot r_N +\tau_{\epsilon}\Big\}\geq P\Big\{d_W\big(P^*,\hat{P}_N\big)\leq r_N\Big\}\geq 1-\beta_N,
\end{align}
where $\beta_N$ is given in \eqref{thm p2}.

Then we denote events $A_N=\Big\{-L\cdot r_N +\frac{\tau_\epsilon}{1+\epsilon} \leq J^* \leq k_{\tau_\epsilon}\cdot r_N +\tau_{\epsilon}\Big\}$. Then $P(A^c_N)\leq\beta_N$. Because $\sum_{N=1}^{\infty}\beta_N < \infty$, we use Borel-Cantelli Lemma and the limit supremum of the sequence of events satisfies:
\begin{align}
    P(\lim_{N\rightarrow\infty} \sup A^c_N) =0.
\end{align}
So $P(\lim_{N\rightarrow\infty} \inf A_N) =1$, which implies the consistency.

Finally, if $\lim_{N\rightarrow \infty}r_N(\beta_N)=0$, we can take $N\rightarrow\infty$  and obtain \eqref{thm p4}. $\hfill\blacksquare$

\subsection{Proof of Theorem \ref{thm:generalization_error}. } 
Utilize  Theorem \ref{thm:confidence_intervals} and Remark \ref{A_N}, we can obtain that:
\begin{align*}
    J^*\leq A_N &\leq k_{\tau_\epsilon}\cdot d_W(P^*, \hat{P}_N)+\tau_\epsilon\\
    &\leq L\cdot d_W(P^*, \hat{P}_N)+\tau_\epsilon\\
    &\leq (2+\epsilon)\cdot L \cdot d_W(P^*, \hat{P}_N)+(1+\epsilon)\cdot J^*,
\end{align*}
 where the second inequality utilizes Lemma \ref{lemma:k tau leq L}. 
 
Then we have the similar step to  \eqref{d to p}:
\begin{align*}
    P\Big\{J^*\leq A_N \leq (1+\epsilon)\cdot J^*+(2+\epsilon)\cdot L \cdot r_N\Big\}\geq P\Big\{d_W\big(P,\hat{P}_N\big)\leq r_N\Big\}\geq 1-\beta_N.
\end{align*}
Then we take the expectation on data and we have:
\begin{align*}
    \bbE_{\sim P_{data}}[d_W(P^*, \hat{P}_N)]&=\int_0^\infty P_{data}\Big\{d_W\big(P,\hat{P}_N\big)\geq r\Big\}dr\\
    &\leq \int_0^1c_1\exp\big(-c_2Nr^{\max\{m,2\}}\big)dr+\int_1^\infty c_1\exp\big(-c_2Nr^a\big)dr\\
    &=N^{-\frac{1}{\max\{m,2\}}} \int_0^{N^{\frac{1}{\max\{m,2\}}}}c_1\exp\big(-c_2t^{\max\{m,2\}}\big)dt+N^{-\frac{1}{a}}\int_{N^{\frac{1}{a}}}^\infty c_1\exp\big(-c_2t^a\big)dt\\
    &\leq N^{-\frac{1}{\max\{m,2\}}} \int_0^{\infty}c_1\exp\big(-c_2t^{\max\{m,2\}}\big)dt+N^{-\frac{1}{a}}\int_0^\infty c_1\exp\big(-c_2t^a\big)dt\\
    &=O(N^{-\min\{\frac{1}{m}, \frac{1}{a}, \frac{1}{2} \}})).
\end{align*}

The first inequality is derived from  \eqref{dw bound} and we utilize the convergence of two exponential integrals. 
Finally, we have:Here $a$ can be removed: When $a>2$ satisfies Assumption \ref{assump:light_tail}, it can be weaken for $a=2$; when $a<2$ in Assumption \ref{assump:light_tail}, here we have $\frac{1}{a}>\frac{1}{2}$, so $\frac{1}{a}$ can be omitted due to minimum. The result should be $O(N^{-\min\{\frac{1}{m},  \frac{1}{2} \}})$.
\begin{align*}
    J^*\leq \bbE_{\sim P_{data}}A_N&\leq (2+\epsilon)\cdot \bbE_{\sim P_{data}}[d_W(P^*, \hat{P}_N)]+(1+\epsilon)\cdot J^*\\
    &=(1+\epsilon)\cdot J^*+O(\frac{1}{N^{\eta}}).
\end{align*} $\hfill\blacksquare$

\subsection{Proof of Theorem \ref{dis. shift}. }
The proof here is very straightforward following the proof of Theorem \ref{thm:confidence_intervals} and \ref{thm:generalization_error}. 
Combine with the formula \eqref{proof chain} that was proven earlier, we can easily obtain similar result:
\begin{align}\label{appendix proof chain}
    -L\cdot d_W(\tilde{P},\hat{P}_N)+\frac{\tau_\epsilon}{1+\epsilon} \leq \Tilde{J} \leq \bbE_{\Tilde{P}}[h(\hat{x}_N,\xi)]\leq k_{\tau_\epsilon}\cdot d_W(\Tilde{P},\hat{P}_N)+\tau_{\epsilon}.
\end{align}
Then we use the triangle inequality of distance: $d_W(\tilde{P},\hat{P}_N)\leq d_W(P^*,\hat{P}_N)+d_W(\tilde{P},P^*)$ and we get the extra term $d_W(\tilde{P},P^*)$. And for the term $d_W(P^*,\hat{P}_N)$, continue to use the tail probability \eqref{dw bound} to get guarantees for various probabilities and expected values. 
$\hfill\blacksquare$

\section{Supplementary results for Section \ref{sec:numerical}}

\subsection{Derivation of Equivalent Models in \cref{sec: RS performance on sample}}\label{appendix derivation of equaivalent models}

\textbf{Proof of \eqref{eq:dro_equivalent}.}   For convenience, let $x_a$ denote the augmented parameter vector $(-x, 1)^T$. Consider Lipschitz loss $h(x,\xi)=L(x_a\cdot \xi)$. For DRO, under mild assumptions, \citeauthor{esfahani2015data} have shown that:
\begin{equation}\label{DRO equivalent form}
    \max_{P\in \{P:d(P,\hat{P}_N)\leq r\}} E_P[h(x,\xi)]= \inf_{\lambda\geq 0 }\lambda r +\frac{1}{N}\sum_{i=1}^N\sup_{\xi}(h(x,\xi)-\lambda c(\xi,\xi_i)).
\end{equation}

Next, denote $\Delta= u  - u_i$, Utilizing the proof given by \citeauthor{shafieezadeh2019regularization}, we can obtain:
\begin{align*}
    &\sup_{\xi}(h(x,\xi)-\lambda c(\xi,\xi_i))
    =\sup_{\xi}(L(x_a\cdot \xi)-\lambda c(\xi,\xi_i))\
    =\sup_{u}(L(x\cdot u-y_i)-\lambda ||u-u_i||_2)\ \\
    &= \sup_{\Delta}(L((u_i+\Delta)\cdot x-y_i)-\lambda||\Delta||_2)
    =\Big\{\begin{array}{ll} &L(u_i\cdot x-y_i) \quad if \lambda \geq ||x||_2,\\& +\infty \quad otherwise. 
    \end{array}   
\end{align*}
The second equality here utilizes the definition of our fine-tuned cost function: if $y$ in $\xi$ and $y_i$ in $\xi_i$ are not equal, the distance will become $\infty$, thereby making the entire expression $-\infty$. Therefore, only the distance of the feature variable $u$ is retained. 

Back to \eqref{DRO equivalent form}, we have:
\begin{equation*}
    \inf_{\lambda\geq 0 }\lambda r +\frac{1}{N}\sum_{i=1}^N\sup_{\xi}(h(x,\xi)-\lambda c(\xi,\xi_i)) = \inf_{\lambda \geq ||x||_2} \lambda r +\frac{1}{N}\sum_{i=1}^NL(u_i\cdot x-y_i)
     =  r ||x||_2 +\frac{1}{N}\sum_{i=1}^NL(u_i\cdot x-y_i).
\end{equation*}

Then we have
\begin{equation}
 \min_{x\in\mathcal{X}}\max_{P\in \{P:d(P,\hat{P}_N)\leq r\}} E_P[h(x,\xi)]=\min_x    \frac{1}{N}\sum_{i=1}^{N}L(y_i-u_i\cdot x)+ r\cdot ||x||_2.   
\end{equation}
$\hfill\blacksquare$

\textbf{Proof of \eqref{RS equivalent}.} We have already given the reformulation of the RS model\eqref{eq:rs_reformulated} in  \cref{sec:tractability}. The constraint condition is:
\begin{equation}\label{reformulation appendix}
    \frac{1}{N}\sum_{i=1}^N[\sup_{z\in\Xi}h(x,z)-kc(\xi_i,z)]\leq \tau.
\end{equation}

The proof will follow the proof of \eqref{eq:dro_equivalent}:
\begin{equation*}
    \frac{1}{N}\sum_{i=1}^N[\sup_{z\in\Xi}h(x,z)-kc(\xi_i,z)]=\Big\{\begin{array}{ll} &\frac{1}{N}\sum_{i=1}^NL(u_i\cdot x-y_i) \quad if k \geq ||x||_2,\\& +\infty \quad otherwise. \end{array}
\end{equation*}
Since $\tau$ serves as the upper bound in \eqref{reformulation appendix}, to ensure that the left side of \eqref{reformulation appendix} is not infinite, it is necessary to satisfy $k\geq ||x||_2$.
Therefore, in \eqref{eq:rs_reformulated}, taking the minimum value of $k$ is equivalent to minimizing $||x||_2$ and let $k=\min_{x}||x||_2$, while satisfying the constraint condition $\frac{1}{N}\sum_{i=1}^NL(u_i\cdot x-y_i)\leq \tau$.
$\hfill\blacksquare$

\textbf{Proof of \eqref{eq:RS_equivalent_2}.} 
Following \eqref{RS equivalent}, we can express it in its dual form: 
\begin{equation*}
    \inf_{x}\sup_{\lambda>0}||x||_2+\lambda\cdot(\frac{1}{N}\sum_{ i=1}^{N}L(y_i-u_i\cdot x)-\tau).
\end{equation*}
Notably, this equation represents a convex problem. As long as $\tau >\min_x\frac{1}{N}\sum_{ i=1}^{N}L(y_i-u_i\cdot x)$ which is mentioned in \eqref{tau greater than optimal}, there exists a point in the relative interior, hence the Slater’s strong duality condition holds. Subsequently, to facilitate comparative analysis with DRO, we replace $\lambda$ with $\frac{1}{\lambda}$ to obtain:
$$\inf_{x}\sup_{\lambda>0}\frac{\frac{1}{N}\sum_{ i=1}^{N}L(y_i-u_i\cdot x)+\lambda\cdot||x||_2-\tau}{\lambda}.$$ 
Finally, note that the above equation is convex with respect to $x$ and concave with respect to $\lambda$. According to the Mini-max theorem, we can interchange the order of sup and inf to obtain the desired formula.
$\hfill\blacksquare$

\subsection{Other Equivalent Conclusions}\label{appendix:Other Equivalent Conclusions}
This section answers the question mentioned in the previous footnote. If we still use the $l_2$ norm of the entire vector as the cost function i.e. $c(\xi_1,\xi_2)=||\xi_1-\xi_2||_2$, then the DRO model will be equivalent to:
\begin{equation*}
 \min_{x\in\mathcal{X}}\max_{P\in \{P:d(P,\hat{P}_N)\leq r\}} E_P[h(x,\xi)]=\min_x    \frac{1}{N}\sum_{i=1}^{N}L(y_i-u_i\cdot x)+ r\cdot ||x_a||_2,   
\end{equation*}
where $x_a$ is the augmented vector $(x,-1)^T$.
Similarly, RS model is equilvalent to:

\begin{align*}
\min_x ~&||x_a||_2, \\
\mbox{s.t.}~ &\frac{1}{N}\sum_{ i=1}^{N}L(y_i-u_i\cdot x)\leq \tau. \nonumber
\end{align*}

And we can also write its dual equivalent form as:
\begin{align*}
\sup_{\lambda>0}\inf_{x}\frac{\frac{1}{N}\sum_{ i=1}^{N}L(y_i-u_i\cdot x)+\lambda\cdot||x_a||_2-\tau}{\lambda}.
\end{align*}

Therefore, after modifying the definition of the cost function, the only difference is whether one term in the model is the $l_2$ norm of the parameter $x$ itself or the $l_2$ norm of the augmented vector $(x,-1)^T$ obtained by adding an element $1$.

\subsection{Function $r(\epsilon)$ in Ten Dimensions}
In  \cref{sec: RS performance on sample}, we set the feature variable to be ten-dimensional. As a supplement to \cref{sec:numerical_shift} , we also plot the $\tau-\epsilon$ relationship graph under the ten-dimensional situation of the feature variable.

\begin{figure}
    \centering
    \includegraphics[scale = 0.5]{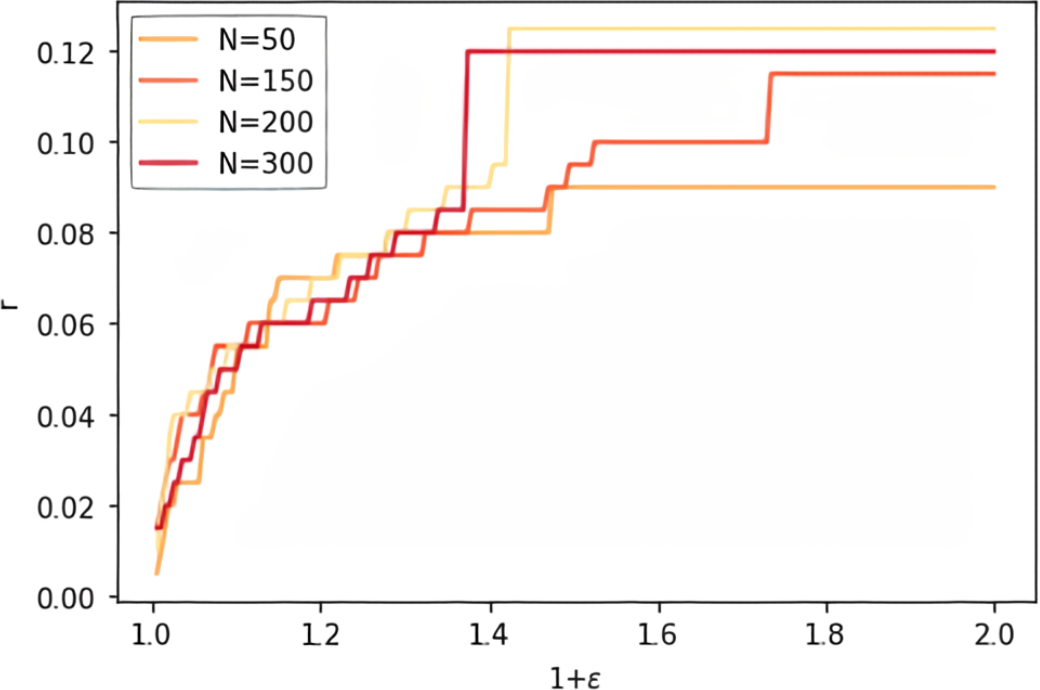}
    \caption{function $r$-$\epsilon(m_u=10)$}
    \label{appendix_r-e_function}
\end{figure}
\cref{appendix_r-e_function} shows similar relationship between the robustness-controlling hyperparameters of the two models: the DRO radius $r$ and the RS tolerance rate $\epsilon$.  The overall trend of the graph is concave. It tends to flatten when $1+\epsilon=1.4$. Moreover, when the function tends to be flat, the corresponding radius value $r$ is smaller than that in the two-dimensional case in \cref{r-e_function}.

\section{Some Classical Loss Functions}\label{sec:ml_loss}
Here we present a few common loss functions.

\begin{table}[H]

\vskip 0.15in
\begin{center}
\begin{small}
\begin{sc}
\resizebox{\textwidth}{!}{
\begin{tabular}{lcccr}
\toprule

          &\textbf{L(z)}&\textbf{Classification(C) or Regression(R)} &\textbf{Learning Model} \\     
\midrule
         \textbf{Hinge Loss} &$\max\{0,1-z\}$ &C&SVM\\ 
         \textbf{Smooth Hinge Loss} &$\Bigg\{\begin{array}{ll}\frac{1}{2}-z & if z\leq0\\ \frac{1}{2}(1-z)^2 &if 0<z<1\\ 0 &z\geq1\end{array} $&C&smooth SVM\\ 
         \textbf{Logloss} &$\log(1+e^{-z})$&C&Logistic Regression\\ 
         \textbf{Squared Loss} &$z^2$&R&MSE\\ 
         \textbf{$L_1$ Loss} &$|z|$&R&MAE\\ 
         \textbf{Huber Loss} &$\Bigg\{\begin{array}{ll}\frac{1}{2}z^2&if |z|\leq \delta\\ \delta(|z|-\frac{1}{2}\delta) &otherwise
         \end{array}$&R&Huber regression\\ 
         \textbf{$\delta$-insensitive Loss} &$\max\{0,|z|-\delta\}$&R& support vector regression\\ 
         \textbf{Pinball Loss} &$\max\{-\delta z, (1-\delta) z\}$&R&Quantile regression\\ 
\bottomrule
\end{tabular}}
\caption{Some Classical Loss Functions}
\label{classical loss functions}
\end{sc}
\end{small}
\end{center}
\vskip -0.1in
\end{table}

Here we consider $h(x,\xi)$ as a loss function for machine learning applications, where $x$ denotes the parameters in the classification or regression model, and $\xi=(\xi^f,\xi^l)^T$ represents the data with $\xi^f$ as the feature variable and $\xi^l$ as the label variable. For binary classification problems, the loss function can be defined as follows:
\begin{align}\label{classification}
    h(x,\xi)=L(\xi^l\cdot x^T \xi^f).
\end{align}

For regression problems, the loss function can be defined as follows:
\begin{align}\label{regression}
    h(x,\xi)=L(\xi^l-x^T\xi^f).
\end{align}

Here we present a few common loss functions (See 
 Table \ref{classical loss functions}).  Apart from the squared loss, all other loss functions in
Table \ref{classical loss functions} are Lipschitz,  so our Assumption \ref{assump:lips} is relatively weak and reasonable. Furthermore, in practical applications, $x$ and $\xi$ are often bounded, so even if we use squared loss, it is Lipschitz in the case of a bounded domain.

\section{Discussion on Lipschitz continuity Assumption}\label{sec:lip continuity}

In our paper, we do not need to assume the Lipschitz continuity of the loss function with respect to $x$, but with respect to $\xi$. This assumption follows that of \cite{esfahani2015data}, with the aim of using the inequality in Proposition \ref{dual w-dis}. We however understand that it is a common condition to assume  the Lipschitz continuity of parameter $x$. In response to this, we  provide a conservative answer: at least for the regression problem in Appendix \ref{sec:ml_loss} where $h(x,\xi)=L(\xi^l-x^T\cdot\xi^f)$, if both the random variable space $\Xi$ and the parameter space $\mathcal{X}$ are bounded, then as long as we assume that $L(\cdot)$ is a Lipschitz function, it can be simultaneously derived that $h(x,\xi)$ is Lipschitz with respect to both $x$ and $\xi=(\xi^f,\xi^l)$. In such scenarios, the Lipschitz assumptions for $x$ and $\xi$ hold simultaneously.

\end{document}